\documentclass[10pt,twocolumn,letterpaper]{article}

\usepackage{cvpr}              

\usepackage{graphicx}
\usepackage{amsmath}
\usepackage{amssymb}
\usepackage{booktabs}

\usepackage{multirow}
\usepackage{url}

\usepackage[pagebackref,breaklinks,colorlinks]{hyperref}

\usepackage[accsupp]{axessibility}  
\usepackage[capitalize]{cleveref}
\crefname{section}{Sec.}{Secs.}
\Crefname{section}{Section}{Sections}
\Crefname{table}{Table}{Tables}
\crefname{table}{Tab.}{Tabs.}

\def\cvprPaperID{*****} 
\def\confName{CVPR}
\def\confYear{2023}

\begin{document}

\title{Surveillance Face Presentation Attack Detection Challenge}
\author{
 Hao Fang$^{\rm 1}$, 
 Ajian Liu$^{\rm 2}$,  
 Jun Wan$^{\rm 1, 2}$\thanks{Contact person},
 Sergio Escalera$^{\rm 3}$,
 Hugo Jair Escalante$^{\rm 4}$,
 Zhen Lei$^{\rm 1, 2, 5}$\\
 $^{\rm 1}$School of Artificial Intelligence, University of Chinese Academy of Sciences, Beijing, China\\
$^{\rm 2}$ MAIS, Institute of Automation, Chinese Academy of Sciences, Beijing, China \\
 $^{\rm 3}$ Universitat de Barcelona and Computer Vision Center, Barcelona, Catalonia, Spain\\
 $^{\rm 4}$ JInstituto Nacional de Astrof\'isica, \'Optica y Electr\'onica, Puebla, Mexico \\ 
$^{\rm 5}$ CAIR, HKISI, CAS\\ 
 \tt\footnotesize
 fanghao21@mails.ucas.ac.cn,
 \tt\footnotesize
 ajianliu92@gmail.com, 
 \tt\footnotesize
 jun.wan@ia.ac.cn \\
}
\maketitle


\begin{abstract}
Face Anti-spoofing (FAS) is essential to secure face recognition systems from various physical attacks. However, most of the studies lacked consideration of long-distance scenarios. Specifically, compared with FAS in traditional scenes such as phone unlocking, face payment, and self-service security inspection, FAS in long-distance such as station squares, parks, and self-service supermarkets are equally important, but it has not been sufficiently explored yet. In order to fill this gap in the FAS community, we collect a large-scale Surveillance High-Fidelity Mask (SuHiFiMask). SuHiFiMask contains $10,195$ videos from $101$ subjects of different age groups, which are collected by $7$ mainstream surveillance cameras. Based on this dataset and protocol-$3$ for evaluating the robustness of the algorithm under quality changes, we organized a face presentation attack detection challenge in surveillance scenarios. It attracted 180 teams for the development phase with a total of 37 teams qualifying for the final round. The organization team re-verified and re-ran the submitted code and used the results as the final ranking. In this paper, we present an overview of the challenge, including an introduction to the dataset used, the definition of the protocol, the evaluation metrics, and the announcement of the competition results. Finally, we present the top-ranked algorithms and the research ideas provided by the competition for attack detection in long-range surveillance scenarios.
\end{abstract}

\section{Introduction}
\label{sec:intro}

In recent years, as face recognition systems are widely used, the security of face recognition systems has been increasingly threatened. Therefore, face Presentation  Attack Detection (PAD) techniques are crucial in defending against malicious attacks and securing face recognition systems. With the release of some high-quality 2D attack datasets~\cite{zhang2012face,chingovska2012effectiveness,boulkenafet2017oulu,Liu2018Learning,zhang2020casia,zhang2019dataset,liu2021casia} and 3D mask datasets~\cite{nesli2013spoofing,erdogmus2013spoofing,liu2022contrastive,liu20163d,manjani2017detecting,steiner2016reliable,george2019biometric}, existing work~\cite{shao2019multi,george2019deep,yu2020searching,zhang2020face,liu2020disentangling,yang2021few,chen2021generalizable,liu2021face,ijcai2022p165,li20203dpc,liu2022disentangling} has achieved satisfactory performance in short-distance applications. However, these algorithms rely heavily on the quality of face images. As a result, their performance degrades significantly in long-distance scenarios, hindering the expansion of FAS work to surveillance contexts. With the popularity of remote cameras and the deployment of surveillance networks, face recognition technology~\cite{Kim_2022_CVPR,8600370,zhong2021sface} is gradually expanding to long-distance applications, such as real-time capture, identity verification in security channels, and payment in self-service supermarkets. Therefore, in order to safeguard society and citizens, we require vision technology that can effectively operate in long-distance surveillance scenarios.

However, the FAS community has conducted little research on deceptive face detection in long-distance surveillance environments. We identify two reasons hindering the development of PAD technology: (1) Lack of surveillance scenario-based datasets. Due to the difficulty of building surveillance scenes and expensive acquisition costs, the existing FAS datasets require subjects to face the acquisition device at a certain distance. However, diverse surveillance scenes, realistic weather and lighting, and natural human behavior are important data features of the remote surveillance PAD dataset. (2) Lack of a common benchmark based on remote surveillance dataset to compare the performance of different algorithms. Current research in the FAS community, both algorithms based on color texture feature learning~\cite{yu2020fas,jia2020survey,jia20203d,george2019biometric} and remote photoplethysmography (rPPG)-based detection~\cite{liu2018remote,lin2019face,liu2021multi} require-high quality face images to extract fine-grained features. However, data based on long-distance surveillance scenarios bring new challenges to FAS algorithms such as low resolution, motion blur, and occlusion. Therefore, a common benchmark based on remote surveillance datasets to drive algorithm advancement is critical.

In order to facilitate the FAS community's research on attack detection in long-distance surveillance scenarios, we targeted the following three aspects to address the current difficulties based on the above analysis: (1) We collected a large-scale PAD dataset based on surveillance scenarios, namely SuHiFiMask~\cite{fang2023surveillance}. Compared to existing short-distance datasets, it has several advantages, such as diverse remote surveillance scenarios, realistic distribution of human faces, a rich pool of spoofing attacks, realistic illumination and diverse weather. (2) We define a more general test protocol for real-world FAS algorithm deployment and publish a public benchmark based on remote surveillance scenarios. We design protocols to evaluate the generality of the algorithms for ID changes, mask type changes, and picture quality changes. (3) Based on the dataset and protocol, we successfully held the \textbf{\emph{Surveillance Face Presentation Attack Detection Challenge at CVPR2023}}\footnote{\url{https://sites.google.com/view/face-anti-spoofing-challenge/welcome/challengecvpr2023?authuser=0}}, which attracted 180 teams from all over the world and significantly boosted the community's research on FAS tasks in surveillance scenarios. A summary of the team names and affiliations that made it to the final stage is shown in Tab~\ref{Table:Team}, We found that most of the teams that qualified for the final round of this competition used large networks as the backbone for their algorithms compared to the previous challenge~\cite{boulkenafet2017competition,liu2019multi,liu2021cross,liu20213d}, suggesting that large models are gradually outperforming traditional vision models on downstream tasks.

\begin{table}[ht]
\footnotesize
\renewcommand\arraystretch{1}
\centering
\caption{Team and affiliations are listed in the final ranking of this challenge.}

\setlength{\tabcolsep}{4mm}{
\scalebox{1}{
\begin{tabular}{|c|c|c|}
\hline
Ranking & Team Name  & Leader Name, Affiliation                                                                    \\ \hline \hline
1       & MateoH     & Keyao Wang, Baidu Inc                                                                       \\ \hline
2       & CTEL\_AI   & Yaowen Xu, China Telecom                                                                    \\ \hline
3       & horsego    & \begin{tabular}[c]{@{}c@{}}Dingheng Zeng, MaShang- \\ Consumer Finance Co., Ltd\end{tabular} \\ \hline
4       & hexianhua  & Xianhua He, Meituan                                                                         \\ \hline
5       & OPDAI      & Heng Cong, NetEase Inc                                                                      \\ \hline
6       & SeaRecluse & Minzhe huang, akuvox                                                                        \\ \hline
7       & XiangR     & Xiang Rao, Xiamen University                                                                \\ \hline
8       & Chenyifan  & TengTeng Zhang, CMB                                                                         \\ \hline
9       & ioNetworks & Yingyu Chen, ioNetworks                                                                     \\ \hline
\end{tabular}
}}
\label{Table:Team}
\end{table}

\section{Challenge Overview}
In this section, we review the organized challenge, including a brief introduction to the surveillance-based dataset SuHiFiMask, an introduction to the protocol produced for this challenge, the process and timeline of the challenge, and the evaluation metrics.

\subsection{SuHiFiMask Dataset}
To the best of our knowledge, SuHiFiMask is the first dataset to extend FAS to real surveillance scenes rather than mimicking low-resolution images and surveillance environments. It contains 10,195 videos, which were taken from 101 subjects of different age groups. For the attack pool, it contains $232$ 3D high-fidelity masks (plaster, resin, silicone, and head mold), $200$ 2D attacks (posters, portraits, and screens), and $2$ adversarial attacks (adversarial hats, adversarial masks). During the collection process, it considers 40 surveillance scenes for video recording, including daily life scenes (\eg, cafes, cinemas, and theaters) and security check scenes (\eg, security check lanes and parking lots) for deploying face recognition systems. In addition, SuHiFiMask takes into account four different types of weather (Sunny, Windy, Cloudy, and Snowy days) and natural lighting (Day and Night light) to imitate the complex and changing real environment. Finally, seven mainstream surveillance cameras were used to record video simultaneously to ensure complete capture of the entire scene.

In order to facilitate the use of the dataset by the participating teams, we performed the following pre-processing: (1) We used RetinaFace~\cite{deng2020retinaface} to detect the face in each frame of the original video. (2) We use face similarity to track the position of faces in consecutive frames. (3) We sample each video in 10-frame intervals and store the cropped face image in the corresponding face tracking box folder. (4) We named the folder of this video according to the following rule: $Group\_Scene\_Camera\_Epoch\_Time$.

\subsection{Challenge Protocol and Data Statistics}
In order to promote the extension of the FAS task to long-range scenarios and to meet the deployment requirements of realistic surveillance networks, we designed a protocol that can evaluate the generalization ability of the algorithm under image quality variations. 
The inference phase of the algorithm degrades in performance due to the input of low-quality images containing noise, which indicates the inadequate generalization ability of the algorithm. Therefore, to better evaluate the generalization ability of the algorithm under quality variations, we constitute the training set, development set and testing set with images of different quality distributions. In detail, we use SER-FIQ~\cite{terhorst2020ser} algorithm to calculate the image quality score which ranges from 0 to 1. As shown in the fifth column of Tab.~\ref{Table:protocol}, we assigned images with scores [0.4, 1] as the training set, scores [0.3, 0.4) as the development set, and scores [0, 0.3) as the testing set. In addition, to accurately evaluate the robustness of the algorithm to quality changes without being affected by other factors, all mask types (referred to as 1,2,3,4), subjects, scenes, weather (Sunny, Windy, Cloudy and Snowy days), and lighting (Day and Night light) were included in the training set, the development set, and the testing set.

\begin{table}[ht]
\footnotesize
\renewcommand\arraystretch{1.5}
\centering
\caption{Statistical information for each protocol of the proposed SuHiFiMask dataset. Note that 1, 2, and 3 in the fourth column mean resin, silicone, and plaster. 4 represents headgear and head mold.}
\setlength{\tabcolsep}{0.7mm}{
\scalebox{1}{
\begin{tabular}{|c|c|c|c|c|c|c|c|c|}
\hline
\multicolumn{1}{|l|}{Prot.} & SubSet & \#Subj. & Mask       & Score & \#Live & \#Mask & \#Others & \#All   \\ \hline \hline
\multirow{3}{*}{3}          & Train  & 101       & 1$\sim$4 & {[}0.4,1{]}   & 64,276 & 35,898 & 58,889         & 159,063 \\ \cline{2-9} 
                            & Dev    & 101       & 1$\sim$4 & {[}0.3,0.4)   & 37,990 & 24,031 & 27,255         & 89,276  \\ \cline{2-9} 
                            & Test   & 101       & 1$\sim$4  & {[}0,0.3)     & 83,181 & 43,087 & 35,614         & 161882  \\ \hline
\end{tabular}
}}
\label{Table:protocol}
\end{table}

\subsection{Challenge Process and Timeline}
The challenge was run in the CodaLab\footnote{\url{https://codalab.lisn.upsaclay.fr/competitions/10080}} platform, and this section will introduce the two stages of this challenge.

\subsubsection{Development Phase} (\emph{Started: February. 1, 2023 - Ended: February 24, 2023}) During this phase, teams have access to labeled training data and unlabeled development data. Participants could use the training data to train their models and submit predictions on the development data. The training data and development data contain all subjects, attack types, surveillance scenarios, weather, lights, and sensors, the only difference is the quality of the images in the training and development sets. In addition, labels for the development set are not provided to participants, who could submit predictions for development data and receive real-time ranking feedback via the leaderboard.

\subsubsection{Final Phase} (\emph{Started: February. 24, 2023 - Ended: March 5, 2023}) In the final phase, the labels of the development set are provided to the teams so that the teams incorporate the development data into the training phase of the model. At the same time, testing sets without labels were published and participants had to make predictions on the test sets and upload the results of their algorithms to the challenge platform. During the final phase, teams will have two opportunities per day to submit their solutions to the challenge platform for the purpose of assessing the stability of their algorithms. Note that the CodaLab platform will display the results of the last submission by default on the challenge website. In addition, to ensure that the final ranking is authentic, teams must publicly release their code under the license of their choice and provide a fact sheet describing their solution.

\subsection{Evaluation Metrics}
For the performance evaluation, we selected the recently standardized ISO/ IEC 30107-3 metrics: Attack Presentation Classification Error Rate (APCER), Normal/ Bona Fide Presentation Classification Error Rate (NPCER/ BPCER) and Average Classification Error Rate (ACER) as the evaluation metric, in which APCER and BPCER/ NPCER are used to measure the error rate of fake or live samples, respectively. The ACER on the testing set is determined by the Equal Error Rate (EER) thresholds on the development set. In addition, the Area Under Curve (AUC) is defined as the area under the ROC curve and its value generally ranges from 0.5 to 1. In this competition, AUC was used as an additional evaluation criterion. Finally, the final ranking of this challenge was determined by the value ACER.

\section{Description of solutions}
\subsection{MateoH}
\begin{figure}[ht]
	\centering
	\includegraphics[width=1.0\linewidth]{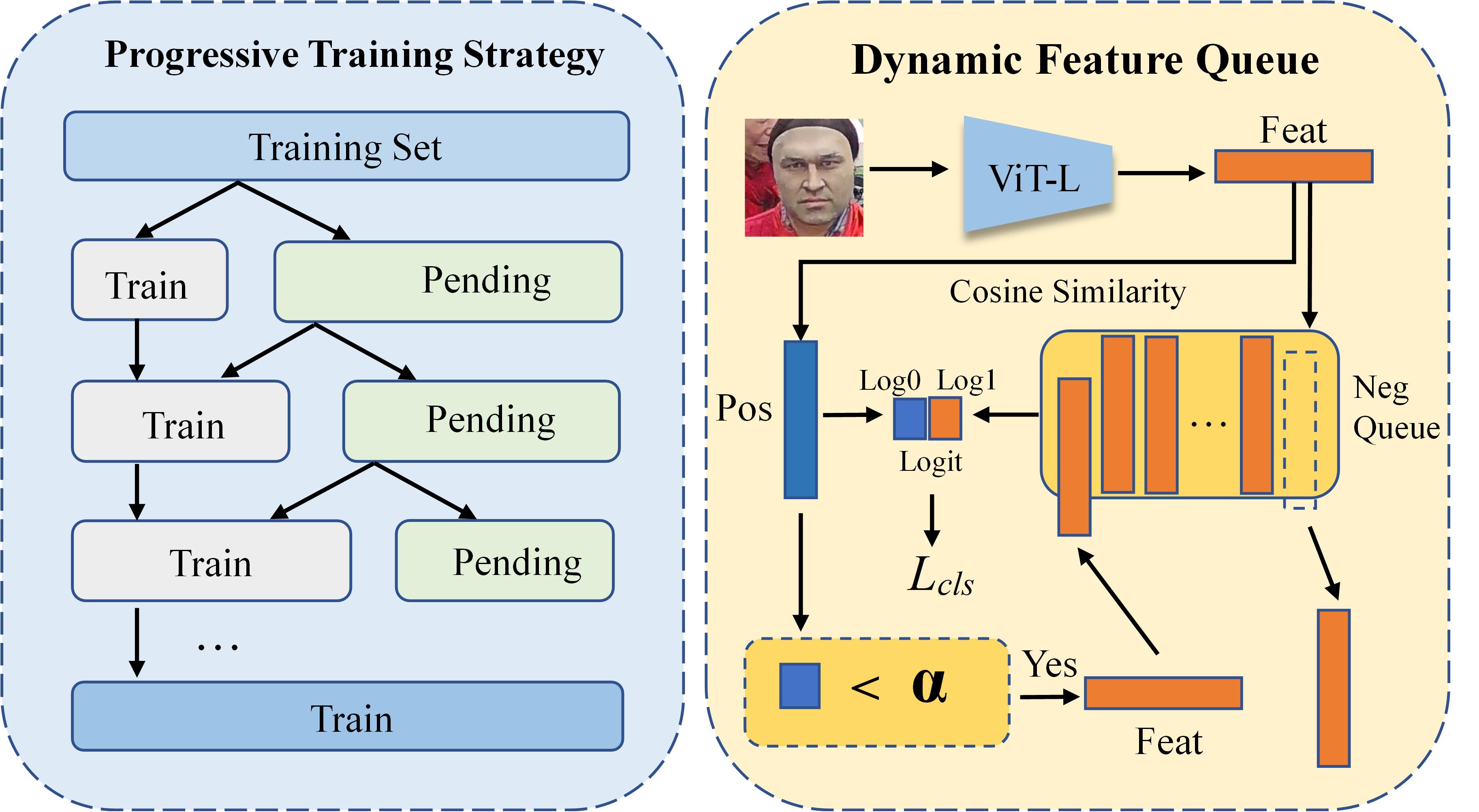}
	\caption{Progressive Training Strategy (PTS) and Dynamic Feature Queue (DFQ) module proposed by team of MateoH. The left side of the figure shows that PTS divides the original training set into a training subset and a pending set, and gradually adds hard samples from the pending set to the training subset. The right side of the figure shows the training process of the DFQ module.}
	\label{fig:1th}
 \end{figure}

Considering that there are many indistinguishable samples in the protocol, the team of MateoH proposes a progressive training strategy (PTS) to solve the catastrophic forgetting problem brought by non-stationary data to the deep network. As shown in Fig.\ref{fig:1th}, the process of PTS is as follows: (1) Sample the initial set of samples. PTS will randomly select a balanced number of positive and negative samples from the training set as the initial training sample set based on the initial sampling rate. The remaining samples are put into the pending set for validation. (2) Hard sample mining. After each training round, the samples in the pending set are predicted with the model and ranked according to their prediction scores. Positive samples with a low score and negative samples with a high score will be added to the training set according to the sampling rate. (3) Reduced sampling rate. A decay factor is used to reduce the sampling rate after each round as a way to gradually increase the focus on hard samples. Overall, the PTS training strategy provides a framework paradigm for progressive hard-sample mining. The strategy allows the model to progressively strengthen its focus on hard samples while playing back previous simple samples to avoid catastrophic forgetting.

To address the issue of t helimited defense capability of FAS
systems against unknown attack types in surveillance scenarios, the team of MateoH proposes a Dynamic Feature Queue (DFQ) module. This module views the negative sample set as a cluster of multiple unknown categories and the positive sample set as a closed sample set. Thus, the binary classification task in FAS is modeled as a multi-classification task thus improving the model's detection of multiple attack types. The training flow of this module is shown in Fig.~\ref{fig:1th}: (1) ViT-Large is used as a backbone network to extract image features. (2) Calculate the cosine similarity log0 of the feature to the negative sample center and the highest cosine similarity log1 of the feature to the negative sample center queue. (3) Concatenate log0 and log1 as logit and calculate the cross entropy loss of logit and label. (4) Judge the size of log0 and threshold $\alpha$, and add the negative samples with log0 less than the threshold to the negative sample queue. (5) Remove the queue head element from the queue and update the network parameters.

\subsection{CTEL\_AI}
\begin{figure}[ht]
	\centering
	\includegraphics[width=1.0\linewidth]{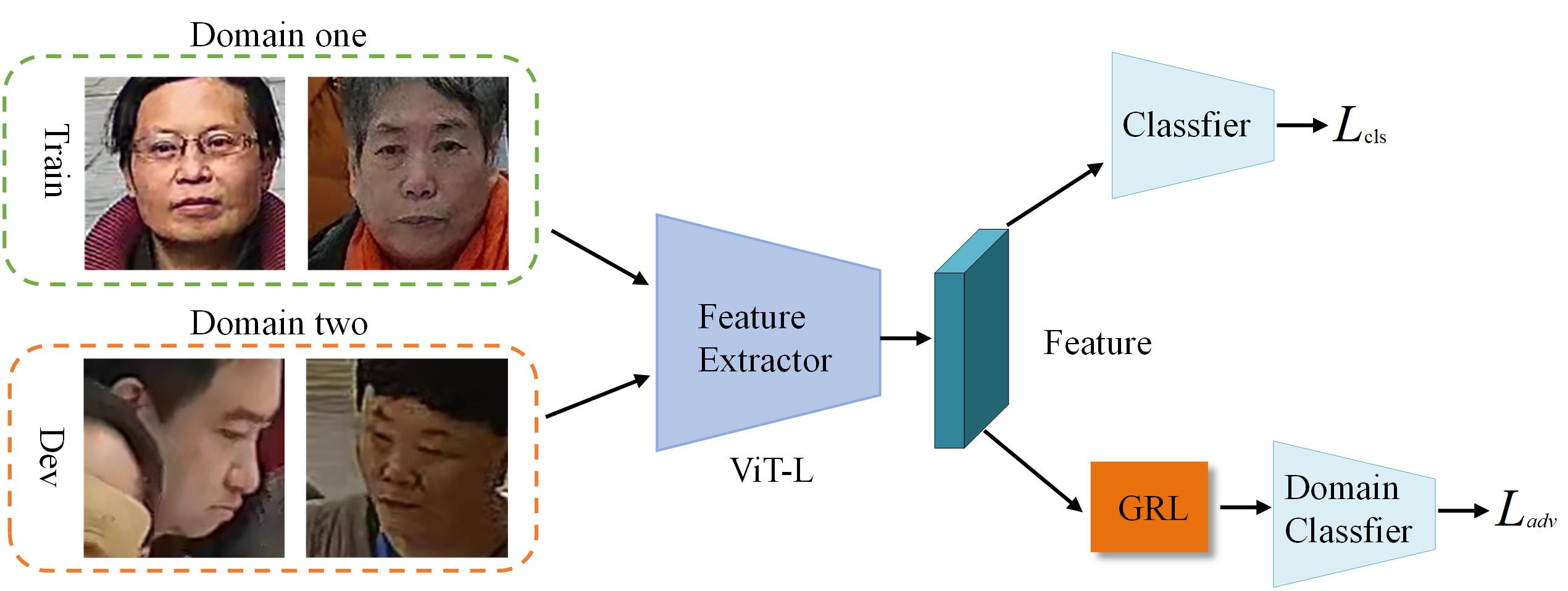}
	\caption{The framework of team CTEL\_AI. The training set and the development set are divided into two different domains. After that, ViT-Large model is used to extract the features of the input images. Cross-entropy loss and adversarial loss are used to optimize the encoder to guide the encoder to extract quality consistent features.}
	\label{fig:2th}
 \end{figure}

Since the training, development, and testing sets for this challenge have different image quality distributions, models with insufficient generalization capabilities will experience degraded performance under the influence of quality differences. To address this problem, the CTEL\_AI team proposes an adversarial domain generalization method that learns a consistent distribution of deceptive features to avoid models that rely too much on detailed features of high-quality images. Specifically, they delineate the domain using the quality score as the boundary, and consider the training set with the picture score interval of [0.4, 1] as domain one and the development set with the picture quality score interval of [0.3, 0.4) as domain two. By optimizing the adversarial domain loss, the encoder learns to extract consistent quality-independent deceptive features. As shown in Fig.~\ref{fig:2th}, the CTEL\_AI team uses the ViT-Large model~\cite{dosovitskiy2020image} as the encoder of the solution and adds the gradient reversal layer GRL~\cite{ganin2015unsupervised} after the encoder to invert the gradient, thus allowing the parameters of the domain classifier to be optimized in the opposite direction.

 In terms of training strategies, SGD was used for gradient descent with a momentum of 0.9 and an initial learning rate (lr) of 0.01. Cosine learning rate decay was employed and the lowest lr is the 1\% of initial lr. An epoch was warm-up with a learning rate, and a total of 100 epochs were trained. It is worth mentioning that based on the above training strategy, they fine-tune the model for a total of 30 epochs after completing 100 epochs of training. In the fine-tuning phase, they used different data augmentation methods to expand the amount of data and gradually added it to the training set. The data augmentation methods include Coarse dropout on the image, Random flip, Random rotate, Random crop, Split patch and shuffle, Gray-scale spoof image, Motion blur, Gaussian blur, Sharpen, Down-sampling, Randomly erase and repeat patches, Random fog, Posterize. The data enhanced by different methods are gradually added to the training set to avoid conflicts caused by using too many enhancement methods at the same time, and also to verify the effectiveness of each method.
 
 Finally, the loss of the whole scheme consists of the cross-entropy loss of the main classifier and the adversarial loss of the domain classifier, as follows:
 \begin{equation}
 \begin{aligned}
\mathcal{L}_{ {total }}=\mathcal{L}_{{cls}}+\mathcal{L}_{{adv}}
\end{aligned}
\label{eq:th2}
\end{equation}

\subsection{horsego}
\begin{figure}[ht]
	\centering
	\includegraphics[width=1.0\linewidth]{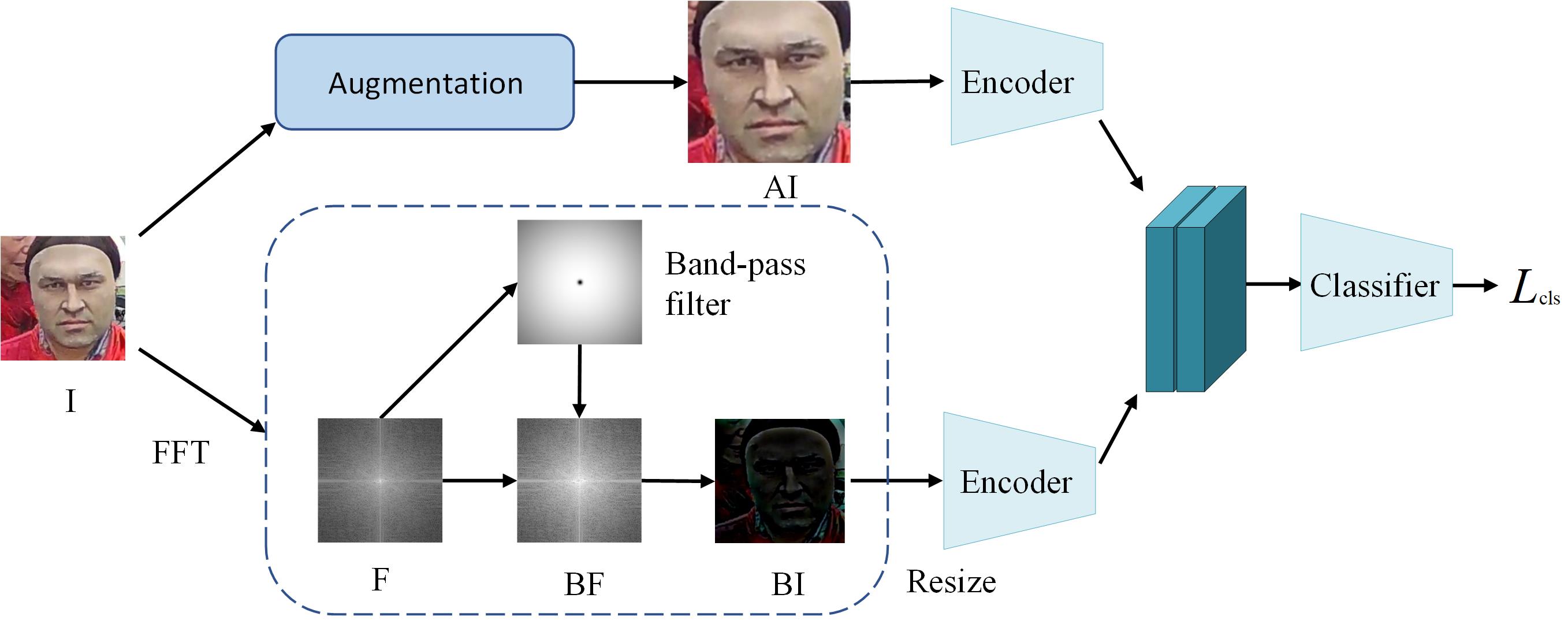}
	\caption{The framework of the team horsego. The AI and BI are obtained by data enhancement and band-pass filtering of the original image I. The two encoders then extracted the features of AI and BI separately and concatenated the features. Finally, the fused features are sent to the main classifier and the cross-entropy loss is used for the optimization of the network.}
	\label{fig:3th}
 \end{figure}

Compared to high-quality images, the training set images of this competition have low resolution, and thus there is little discriminative information in the images for the network to learn. To solve this problem, the team of horsego filtered the original images through a band-pass filter to obtain additional frequency domain information. In addition, they concatenate the spatial domain feature and frequency domain feature of the images before the fully connected layer to fuse them across channels, which makes the binary classification results more accurate. As shown in Fig.~\ref{fig:3th}, they use a dual-stream structure, where one branch performs data enhancement and transformation on the original image to obtain a three-channel image of size 224x224. The other branch uses a band-pass filter to mask some high-frequency information and low-frequency information of the original image, and then converts the image into a filtered map of size 224x224 by inverse fast Fourier transform. Then, two EfficienetFormerV2~\cite{li2022rethinking} are used as encoders to extract frequency domain features and spatial domain features of the processed images, respectively. Finally, these two features are fused across channels and fed to the main classifier for classification, and the cross-entropy loss is used to supervise the main classifier.

In detail, the steps for data pre-processing through band-pass filters are as follows: (a) The original image undergoes Fast Fourier Transform (FFT) transformation and translation to obtain a frequency spectrum F with energy centralized. (b) Two Gaussian kernels with different variances are subtracted to obtain the band-pass kernel K. (c)  The point product of F and K is calculated to obtain the frequency domain band-pass filtered frequency spectrum BF. (d) BF undergoes translation, IFFT (Inverse Fast Fourier Transform) transformation, and is taken as a real value to obtain the band-pass filtering image (BI). In addition, Google's newly proposed LION optimizer~\cite{chen2023symbolic} is used for network training, which has faster convergence and higher accuracy compared to other optimizers.

\subsection{hexianhua}
 \begin{figure*}[ht]
	\centering
	\includegraphics[width=1.0\linewidth,height=0.25\textwidth]{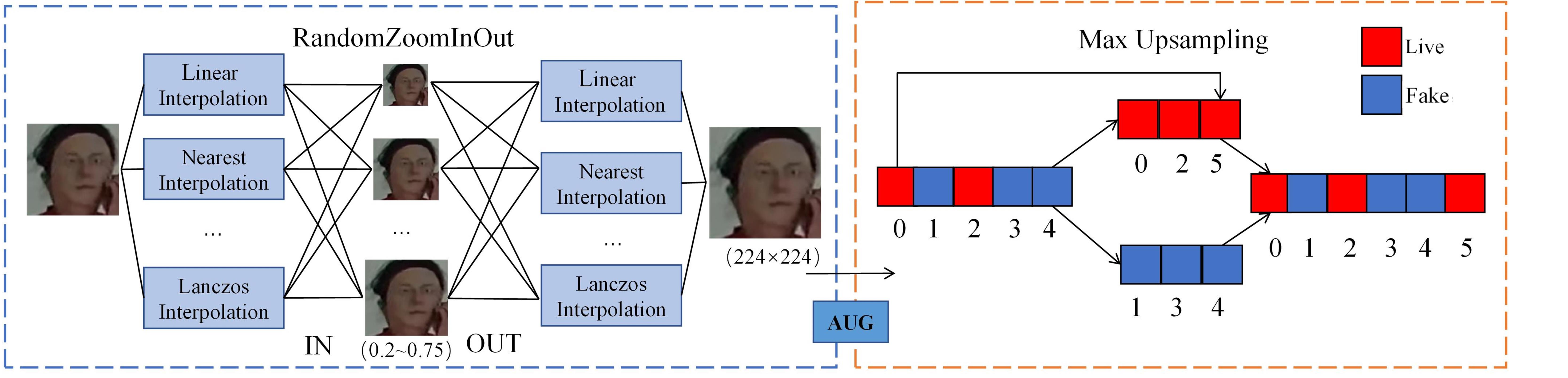}
	\caption{Pre-processing strategies used by the team of hexianhua. RandomZoomInOut data enhancement method randomly picks an interpolation method to shrink the image. The Max Upsampling method mitigates the long tail problem by complementing fewer categories. The AUG represents the various methods.}
	\label{fig:4th}
 \end{figure*}
In the acquisition process of the challenge dataset, multiple surveillance cameras capture the same face simultaneously from different distances and angles. The team of hexianhua refers to such images with different perspectives of the same identity as homologous images. They propose a new data pre-processing strategy to effectively simulate homologous images of different quality to expand the training set, so that the model learns features of different quality distributions. As shown in Fig.~\ref{fig:4th}, in the RandomZoomInOut module, the original images are randomly reduced to between 0.20x and 0.75x, and the interpolation algorithms used for the reduction (Nearest neighbor interpolation, Linear interpolation, Area interpolation, Lancozos interpolation) are also chosen randomly. Then, the interpolation algorithms are again randomly selected to resize these low-resolution images to 224$\times$224 to fit the input size of the encoder.

 Moreover, considering that the number of live samples and the number of attack samples in the training samples are not balanced, the team of hexianhua uses the data balancing strategy of Max Upsampling to mitigate the performance degradation caused by the long tail problem. As shown in the right side of Fig.~\ref{fig:4th}, they use the total number of categories with more samples as the number of Up-sampling. Then they repeatedly selected images from the set of images in the fewer categories to add to the training set until the number of images in the two categories in the training set is equal. In addition, they use the training method of EMA (Exponential Moving Average) and the data post-processing method of TTA (Test Time Augmentation) to further improve the performance of the model. 
 
 Finally, SwinV2-Huge~\cite{liu2021swin} is used as the backbone network of the method, and both cross-entropy loss and Focal loss~\cite{lin2017focal} are used to constrain the encoder, making it give more attention to some hard samples while learning discriminative features. The total loss function formulates as follows:
  \begin{equation}
 \begin{aligned}
\mathcal{L}_{ {total }}=\mathcal{L}_{{cls}}+0.5 *\mathcal{L}_{{focal}}
\end{aligned}
\label{eq:th4}
\end{equation}

\subsection{OPDAI}
The team of OPADAI proposed a dual-stream network, with ConvNextv2\_base and ConvNeXtv2\_large~\cite{woo2023convnext} as the backbone of the two branches. They also added an MLP for each branch to improve the representativeness of the features extracted from different branches by optimizing the respective classifier parameters. Finally, they concatenate the features extracted by the two extractors and classify the concatenated features with an independent MLP, which makes the features learned by the two encoders more effective after fusion. As shown in Fig.~\ref{fig:5th}, three MLP layers are supervised with focal loss to enhance the model's focus on difficult samples in surveillance scenarios. Where loss1 is the focal loss of MLP1 for the classification of fused features, and loss2 and loss3 are the focal losses of MLP2 and MLP3 for the classification of branching features, respectively. The total loss function is expressed as follows:
 \begin{figure}[ht]
	\centering
	\includegraphics[width=1.0\linewidth]{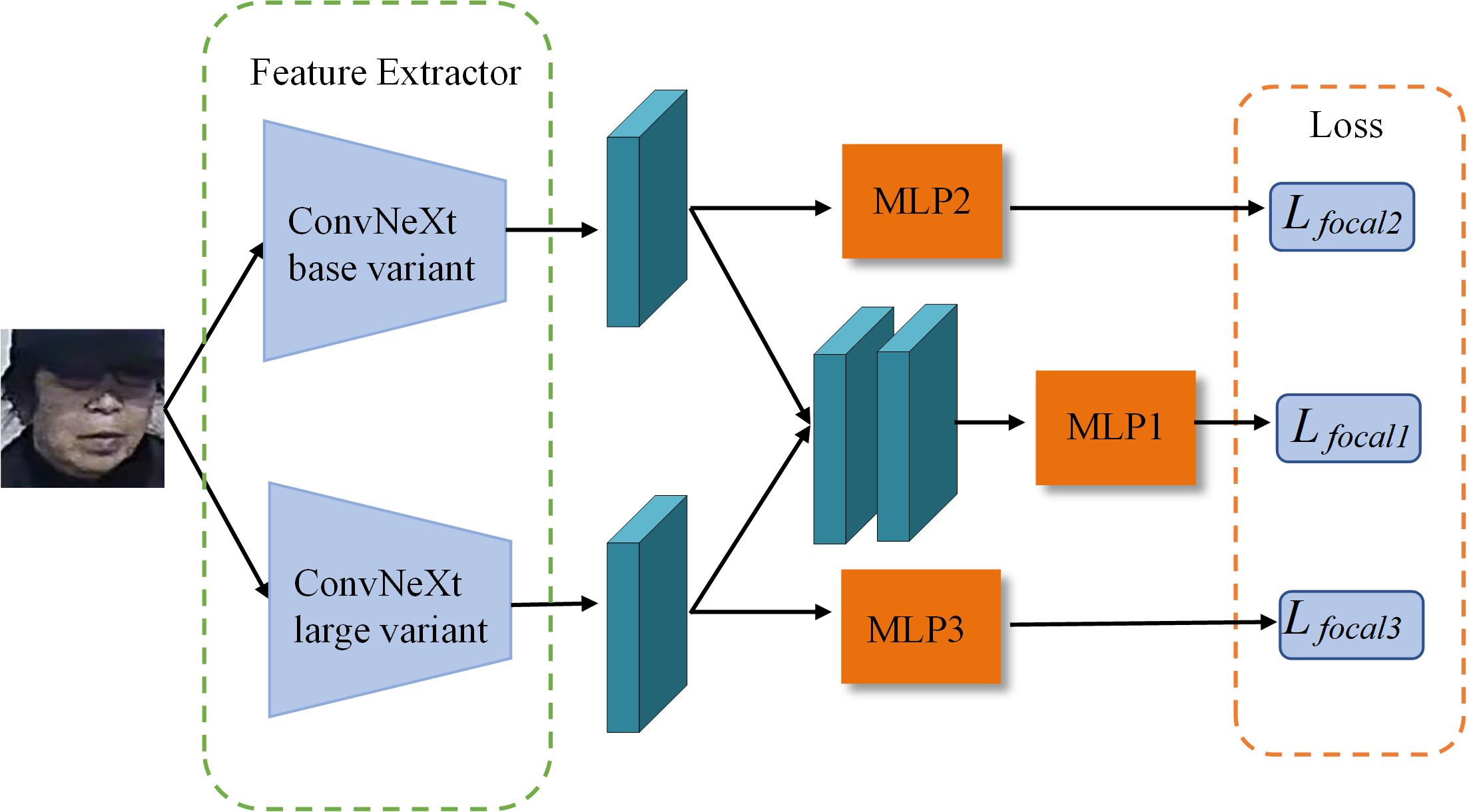}
	\caption{The framework of team OPDAI. The two branches extract features with different variants of ConvNextv2 respectively, and subsequently, the two features are concatenated as fused features. Finally, the three features are sent to separate MLP modules and supervised with three focal losses.}
	\label{fig:5th}
 \end{figure}
 
  \begin{equation}
 \begin{aligned}
\mathcal{L}_{ {total }}=\mathcal{L}_{{focal1}}+0.5 *\mathcal{L}_{{focal2}}+0.5 *\mathcal{L}_{{focal3}}
\end{aligned}
\label{eq:th5}
\end{equation}

In terms of training strategy, they contribute a scheme to optimize the overall network framework. First, they fixed the parameters of the two feature extractors and pre-trained the three MLPs with a batch size of 512. Then they fine-tuned the whole dual-stream network with a batch size of 200, and the training process continued until the sum of the losses dropped to 0.0001. This has the following two advantages: (1) Allowing the fully connected layer in the MLP to obtain the appropriate parameters after several iterations helps to quickly mine the hard samples in the dataset that are difficult to discriminate. (2) Fine-tuning the whole network framework after pre-training enables the encoder to retain the ability to classify easy samples. In addition, they experimentally demonstrate that the AdamW optimizer and TTA (test time augmentation) can achieve superior performance when dealing with complex and variable images.

\subsection{SeaRecluse}
Since the challenge datasets contain multiple sizes of images, the team of SeaRecluse tries to find a uniform size of the input images that can maximize the performance of the algorithm. Specifically, they trained multiple models simultaneously, and during the training of each different model, they scaled the data in advance to a fixed size such as 112$\times$112, 128$\times$128, 168$\times$168, 192$\times$192, 224$\times$224. Then, they tested the trained model on the validation set and recorded the score for each test image. During the statistical process, they found that the model trained on data of size 224$\times$224 had the best generalization performance for the images in the surveillance scenario.

In general, they used  ConvNeXt\_base~\cite{liu2022convnet} as the backbone network of their method and used the techniques of mix up~\cite{zhang2017mixup} and label smoothing~\cite{szegedy2016rethinking} to improve the generalization of the network to complex data distributions. For the training strategy, they choose SoftTargetCrossEntropy as the loss function and use the CyclicLRd scheduler to let the learning rate cycle between 1e-5 and 0.002. It is worth mentioning the data enhancement scheme considering that image flipping was used in the training phase. During the model inference, they weighted the prediction scores of the original and flipped images as the final prediction result. 

\subsection{XiangR}
In order to explore the association between model size and algorithm performance, the XiangR team at Xiamen University trained ConvNext variants with different parameter scales (ConvNext-T, ConvNext-B, ConvNext-L, ConvNext-XL) under the same experimental setup. They found that the larger model performed better on the dataset of this challenge, but constrained by hardware resources and time constraints, they chose ConvNext-B as the solution backbone network.

Before starting the training, they used various pre-processing methods such as Random Resized Crop, Horizontal Flip, and Random Affine to increase the diversity of the data. Then, he used Mix-up, Transfer Learning, Label Smoothing, and other tricks to improve the model's ability to generalize data with the complex distribution. For the training strategy, they chose the AdamW optimizer, which is initialized with a learning rate of 0.01 and a weight decay factor of 1E-2. In addition, a cosine annealing learning rate scheduler is used, which gradually decreases the learning rate in a cosine pattern. This scheduler has a maximum number of epochs of 100 and a minimum learning rate of 0. In other words, the AdamW optimizer with weight decay helps prevent over-fitting, while the cosine annealing scheduler gradually decreases the learning rate to accelerate the model optimization. Finally, the cross-entropy loss is used to supervise the training process of the model.

\subsection{chenyifan} 
\begin{figure}[ht]
	\centering
	\includegraphics[width=1.0\linewidth, height=0.25\textwidth]{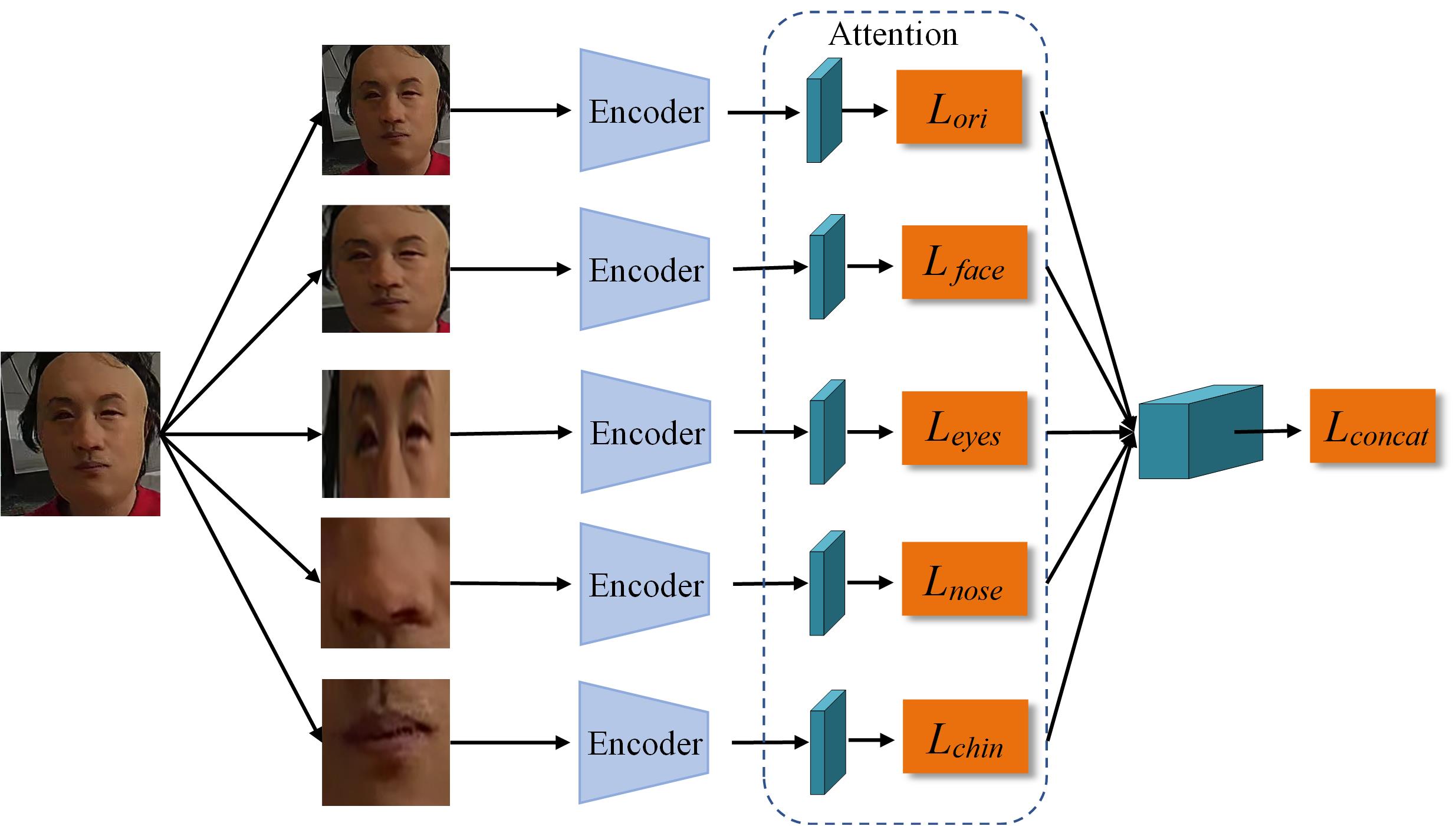}
	\caption{The pipeline of the team chenyifan. The original image and four face patches cropped with prior knowledge are sent to each of the five encoders for feature extraction. The five features are then sent to the attention module and are concatenated as fused features. Finally, to supervise the classification of these six features, six focal losses are employed.}
	\label{fig:8th}
 \end{figure}

Considering that various patches of the face contain deceptive features, the team of chenyifan proposes a high-resolution face patch pipeline based on the original image cropped out, as shown in Fig.~\ref{fig:8th}. The networks on these different branches can focus on learning deceptive traces on a particular face patch, and these traces are used as additional information to assist in the classification of the whole model. In the ready state, the original image is resized to a size of 256$\times$256. The bounding box used to crop individual patches is predefined to be no more than 256$\times$256. The face boundaries are then used to crop four face patches: face, eyes, nose, and chin. It is worth stating that the original image is also used as a branch input. Then, each patch was resized to 224$\times$224 and input to each network branch after mix up~\cite{zhang2017mixup}. As shown in Fig.~\ref{fig:8th}, the team of chenyifan builds a multi-branch network, including branches for the original image, face, eyes, nose, and chin, each branch has a separate encoder to learn specific features, and the five encoders are efficientnetv2\_m, efficientnetv2\_b2,  efficientnetv2\_b1, efficientnetv2\_b0, efficientnetv2\_b0. Next, they input these five features into the attention module and supervise them with binary cross-entropy (BCE) loss. Finally, these five features are concatenated to obtain the fused deception information and perform classification. The final losses are as follows:
  \begin{equation}
 \begin{aligned}
\mathcal{L}_{ {total }}=3*\mathcal{L}_{{ori}}+\mathcal{L}_{{face}}+0.5 *\mathcal{L}_{{eyes}}\\+0.5 *\mathcal{L}_{{nose}}+0.5 *\mathcal{L}_{{chain}}+3 *\mathcal{L}_{{concat}}
\end{aligned}
\label{eq:th8}
\end{equation}
where all the losses are BCE losses. Since patches such as the nose, eyes, and chin contain little deceptive information, they reduce the weights of these loss terms to 0.5 times to reduce the confidence of the whole network in this auxiliary information. For the training strategy, the cosine annealing schedule is used to decay the learning rate, where the initial learning rate is set to 0.0001 and the learning cycle decay is set to 0.5.

\subsection{ioNetworks}
 \begin{figure*}[ht]
	\centering
	\includegraphics[width=1\linewidth,height=0.24\textwidth]{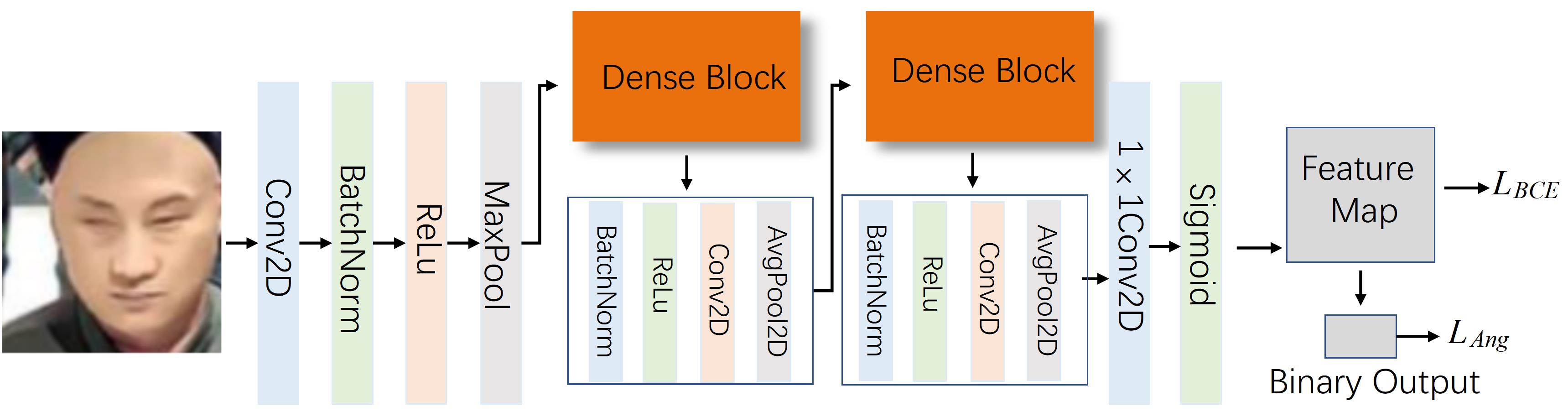}
	\caption{The framework of the team ioNetworks. A DeepPixBis-based framework is used to extract features. Binary cross-entropy loss and sub-center angular margin loss are used simultaneously to optimize the network.}
	\label{fig:9th}
 \end{figure*}
The team of ioNetworks tries to use feature maps from multiple scales to obtain spoof traces. They use DeepPixBis~\cite{george2019deep} as the basic framework, DenseNet as the backbone to reuse features from the previous layers, and fuse feature maps from multiple scales. 

They utilize BCE to supervise the output feature maps and encourage the network to learn local information about the image patches. As shown in Fig.~\ref{fig:9th}, to improve the robustness of the model to noise in the challenge dataset, inspired by advanced face recognition research~\cite{deng2020sub}, they also employ sub-center angular margin loss~\cite{hossain2020deeppixbis} to supervise the binary output. This loss encourages sub-classes containing clean samples to dominate the optimization of the model thus improving the robustness of the model. The overall loss function is as follows:
\begin{equation}
\begin{aligned}
\mathcal{L}_{ {total }}=0.5*\mathcal{L}_{{BCE}}+0.5 *\mathcal{L}_{{Ang}}
\end{aligned}
\label{eq:th9}
\end{equation}
where the second term of this formula is actually a variant of binary cross-entropy loss, which is calculated as:
\begin{equation}
 \begin{aligned}
\mathcal{L}_{{Ang}}=-1/ n\sum_{i}(t[i]*cos(\theta+0.5)\\+(1-t[i]) * \log (1-cos(\theta)))
\end{aligned}
\label{eq:th10}
\end{equation}
where $t[i]$ represents the label of the i-th sample and $theta$ represents the angle between the features learned by the backbone network and the weights.

For the training strategy, they used AdamW as an optimizer, with a learning rate of 0.0001 and a weight decay factor of 0.0005. To adjust the learning rate during training, they utilized a scheduler that decreased the learning rate for each parameter group by a factor of 0.8 every 20 epochs. Finally, they implemented early stopping with patience of 50 epochs to halt model training at a better convergence time.

\section{Challenge Results}
\subsection{Challenge Results Report }
We adopted four metrics to evaluate the performance of the solutions, which are APCER, NPCER, ACER, and AUC respectively. Please note that although we report performance on various evaluation metrics, ACER is the leading metric that determines the final ranking of the participating teams. As shown in Tab.~\ref{Table:result}, which lists the results and ranking of the top 9 teams. We can draw the following three conclusions: (1) The first-place team got the best results on indicators ACER, APCER, AUC, and the second-place team got the best results on BPCER. (2) Eight of the top nine teams are from the industry, indicating that the task of attack detection in surveillance scenarios is important in real-world applications. (3) The top nine teams' ACER scores were distributed between 4\% and 12\% and there was some disparity in the results between teams, which shows that face attack detection under surveillance is challenging and proves the value of SuHiFiMask's in-depth research.
\begin{table}[ht]
\footnotesize
\renewcommand\arraystretch{1.5}
\centering
\caption{Team and results are listed in the final ranking of this challenge.}
\setlength{\tabcolsep}{1.5mm}{
\scalebox{1}{
\begin{tabular}{|c|c|c|c|c|c|}
\hline
\multicolumn{1}{|l|}{R.} & Team       & ACER(\%)      & APCER(\%)     & BPCER(\%)     & AUC(\%)        \\ \hline \hline
1                        & MateoH     & \textbf{4.73} & \textbf{5.07} & 4.38          & \textbf{98.38} \\ \hline
2                        & CTEL\_AI   & 5.56          & 9.20          & \textbf{1.91} & 98.21          \\ \hline
3                        & horsego    & 6.22          & 8.17          & 4.26          & 96.97          \\ \hline
4                        & hexianhua  & 7.08          & 11.21         & 2.94          & 97.83          \\ \hline
5                        & OPDAI      & 7.16          & 9.18          & 5.13          & 97.38          \\ \hline
6                        & SeaRecluse & 9.89          & 15.92         & 3.86          & 96.02          \\ \hline
7                        & XiangR     & 9.96          & 11.35         & 8.57          & 95.80          \\ \hline
8                        & Chenyifan  & 11.01         & 14.12         & 7.89          & 94.39          \\ \hline
9                        & ioNetworks & 12.00         & 15.47         & 8.53          & 95.13          \\ \hline
\end{tabular}
}}
\label{Table:result}
\end{table}

\subsection{Competition summary and Future Work}
Through the challenge, we summarize the effective ideas for surveillance face presentation attack detection: (1) For backbone networks, the performance is superior for networks with a large number of parameters, which proves that large models play an important role in complex applications in real life. (2) At the data level, using pre-processing methods to increase the diversity of data can effectively prevent model over-fitting, and the method of balancing the number of categories can improve the stability of the algorithm. (3) Multi-branching based feature learning method has been shown by several teams to be an effective approach for FAS tasks. In the following work, we further improve the performance from the following aspects: (1) We will explore more efficient super-resolution~\cite{wang2018esrgan,chan2021basicvsr,wang2021towards} methods to recover detailed features of low-quality face images. (2) We will generate~\cite{he2022gcfsr} multiple images with similar data distribution as SuHiFiMask to guide the model to learn more general features. (3) We will explore the interpretability and validity of the large model for FAS tasks.

\section{Conclusion }
We organized the \textbf{\emph{Surveillance Face Presentation Attack Detection Challenge at CVPR2023}} based on the SuHiFiMask dataset and running on the CodaLab platform. 180 teams registered for the competition and 37 teams made it to the final stage. In the final stage of the competition, a total of nine teams submitted their code and participated in the ranking of results after verification by the organizers. We first introduce the associated dataset, the competition protocol, and the evaluation metrics. Then, we review the solutions of the participating ranked teams and report the results of the final phase. Finally, we summarize the conclusions related to the challenges and point out effective methods for attack detection for remote monitoring scenarios.

\section*{Acknowledgments}
This work was supported by the National Key Research and Development Plan under Grant 2021YFF0602103, the External cooperation key project of Chinese Academy Sciences 173211KYSB20200002, the Chinese National Natural Science Foundation Projects 61876179, 61961160704, and 62276254, the Science and Technology Development Fund of Macau Project (No.$\sim $0010/2019/AFJ, 0008/2019/A1, 0025/2019/AKP, 0019/2018/ASC), and the InnoHK program.

{\small
\bibliographystyle{ieee_fullname}
\bibliography{egbib}

\begin{thebibliography}{10}\itemsep=-1pt

\bibitem{boulkenafet2017competition}
Zinelabdine Boulkenafet, Jukka Komulainen, Zahid Akhtar, Azeddine Benlamoudi,
  Djamel Samai, Salah~Eddine Bekhouche, Abdelkrim Ouafi, Fadi Dornaika,
  Abdelmalik Taleb-Ahmed, Le Qin, et~al.
\newblock A competition on generalized software-based face presentation attack
  detection in mobile scenarios.
\newblock In {\em 2017 IEEE International Joint Conference on Biometrics
  (IJCB)}, pages 688--696. IEEE, 2017.

\bibitem{boulkenafet2017oulu}
Zinelabinde Boulkenafet, Jukka Komulainen, Lei Li, Xiaoyi Feng, and Abdenour
  Hadid.
\newblock Oulu-npu: A mobile face presentation attack database with real-world
  variations.
\newblock In {\em 2017 12th IEEE international conference on automatic face \&
  gesture recognition (FG 2017)}, pages 612--618. IEEE, 2017.

\bibitem{chan2021basicvsr}
Kelvin~CK Chan, Xintao Wang, Ke Yu, Chao Dong, and Chen~Change Loy.
\newblock Basicvsr: The search for essential components in video
  super-resolution and beyond.
\newblock In {\em Proceedings of the IEEE/CVF Conference on Computer Vision and
  Pattern Recognition}, pages 4947--4956, 2021.

\bibitem{chen2023symbolic}
Xiangning Chen, Chen Liang, Da Huang, Esteban Real, Kaiyuan Wang, Yao Liu, Hieu
  Pham, Xuanyi Dong, Thang Luong, Cho-Jui Hsieh, et~al.
\newblock Symbolic discovery of optimization algorithms.
\newblock {\em arXiv preprint arXiv:2302.06675}, 2023.

\bibitem{chen2021generalizable}
Zhihong Chen, Taiping Yao, Kekai Sheng, Shouhong Ding, Ying Tai, Jilin Li,
  Feiyue Huang, and Xinyu Jin.
\newblock Generalizable representation learning for mixture domain face
  anti-spoofing.
\newblock {\em arXiv preprint arXiv:2105.02453}, 2021.

\bibitem{chingovska2012effectiveness}
Ivana Chingovska, Andr{\'e} Anjos, and S{\'e}bastien Marcel.
\newblock On the effectiveness of local binary patterns in face anti-spoofing.
\newblock In {\em 2012 BIOSIG-proceedings of the international conference of
  biometrics special interest group (BIOSIG)}, pages 1--7. IEEE, 2012.

\bibitem{deng2020sub}
Jiankang Deng, Jia Guo, Tongliang Liu, Mingming Gong, and Stefanos Zafeiriou.
\newblock Sub-center arcface: Boosting face recognition by large-scale noisy
  web faces.
\newblock In {\em Computer Vision--ECCV 2020: 16th European Conference,
  Glasgow, UK, August 23--28, 2020, Proceedings, Part XI 16}, pages 741--757.
  Springer, 2020.

\bibitem{deng2020retinaface}
Jiankang Deng, Jia Guo, Evangelos Ververas, Irene Kotsia, and Stefanos
  Zafeiriou.
\newblock Retinaface: Single-shot multi-level face localisation in the wild.
\newblock In {\em Proceedings of the IEEE/CVF conference on computer vision and
  pattern recognition}, pages 5203--5212, 2020.

\bibitem{dosovitskiy2020image}
Alexey Dosovitskiy, Lucas Beyer, Alexander Kolesnikov, Dirk Weissenborn,
  Xiaohua Zhai, Thomas Unterthiner, Mostafa Dehghani, Matthias Minderer, Georg
  Heigold, Sylvain Gelly, et~al.
\newblock An image is worth 16x16 words: Transformers for image recognition at
  scale.
\newblock {\em arXiv preprint arXiv:2010.11929}, 2020.

\bibitem{erdogmus2013spoofing}
Nesli Erdogmus and S{\'e}bastien Marcel.
\newblock Spoofing 2d face recognition systems with 3d masks.
\newblock In {\em 2013 International Conference of the BIOSIG Special Interest
  Group (BIOSIG)}, pages 1--8. IEEE, 2013.

\bibitem{fang2023surveillance}
Hao Fang, Ajian Liu, Jun Wan, Sergio Escalera, Chenxu Zhao, Xu Zhang, Stan~Z
  Li, and Zhen Lei.
\newblock Surveillance face anti-spoofing.
\newblock {\em arXiv preprint arXiv:2301.00975}, 2023.

\bibitem{ganin2015unsupervised}
Yaroslav Ganin and Victor Lempitsky.
\newblock Unsupervised domain adaptation by backpropagation.
\newblock In {\em International conference on machine learning}, pages
  1180--1189. PMLR, 2015.

\bibitem{george2019deep}
Anjith George and S{\'e}bastien Marcel.
\newblock Deep pixel-wise binary supervision for face presentation attack
  detection.
\newblock In {\em 2019 International Conference on Biometrics (ICB)}, pages
  1--8. IEEE, 2019.

\bibitem{george2019biometric}
Anjith George, Zohreh Mostaani, David Geissenbuhler, Olegs Nikisins, Andr{\'e}
  Anjos, and S{\'e}bastien Marcel.
\newblock Biometric face presentation attack detection with multi-channel
  convolutional neural network.
\newblock {\em IEEE Transactions on Information Forensics and Security},
  15:42--55, 2019.

\bibitem{he2022gcfsr}
Jingwen He, Wu Shi, Kai Chen, Lean Fu, and Chao Dong.
\newblock Gcfsr: a generative and controllable face super resolution method
  without facial and gan priors.
\newblock In {\em Proceedings of the IEEE/CVF Conference on Computer Vision and
  Pattern Recognition}, pages 1889--1898, 2022.

\bibitem{hossain2020deeppixbis}
Md~Sourave Hossain, Labiba Rupty, Koushik Roy, Md Hasan, Shirshajit Sengupta,
  and Nabeel Mohammed.
\newblock A-deeppixbis: Attentional angular margin for face anti-spoofing.
\newblock In {\em 2020 Digital Image Computing: Techniques and Applications
  (DICTA)}, pages 1--8. IEEE, 2020.

\bibitem{jia2020survey}
Shan Jia, Guodong Guo, and Zhengquan Xu.
\newblock A survey on 3d mask presentation attack detection and
  countermeasures.
\newblock {\em Pattern recognition}, 98:107032, 2020.

\bibitem{jia20203d}
Shan Jia, Xin Li, Chuanbo Hu, Guodong Guo, and Zhengquan Xu.
\newblock 3d face anti-spoofing with factorized bilinear coding.
\newblock {\em IEEE Transactions on Circuits and Systems for Video Technology},
  31(10):4031--4045, 2020.

\bibitem{Kim_2022_CVPR}
Minchul Kim, Anil~K. Jain, and Xiaoming Liu.
\newblock Adaface: Quality adaptive margin for face recognition.
\newblock In {\em Proceedings of the IEEE/CVF Conference on Computer Vision and
  Pattern Recognition (CVPR)}, pages 18750--18759, June 2022.

\bibitem{8600370}
Pei Li, Loreto Prieto, Domingo Mery, and Patrick~J. Flynn.
\newblock On low-resolution face recognition in the wild: Comparisons and new
  techniques.
\newblock {\em IEEE Transactions on Information Forensics and Security},
  14(8):2000--2012, 2019.

\bibitem{li20203dpc}
Xuan Li, Jun Wan, Yi Jin, Ajian Liu, Guodong Guo, and Stan~Z Li.
\newblock 3dpc-net: 3d point cloud network for face anti-spoofing.
\newblock In {\em 2020 IEEE International Joint Conference on Biometrics
  (IJCB)}, pages 1--8. IEEE, 2020.

\bibitem{li2022rethinking}
Yanyu Li, Ju Hu, Yang Wen, Georgios Evangelidis, Kamyar Salahi, Yanzhi Wang,
  Sergey Tulyakov, and Jian Ren.
\newblock Rethinking vision transformers for mobilenet size and speed.
\newblock {\em arXiv preprint arXiv:2212.08059}, 2022.

\bibitem{lin2019face}
Bofan Lin, Xiaobai Li, Zitong Yu, and Guoying Zhao.
\newblock Face liveness detection by rppg features and contextual patch-based
  cnn.
\newblock In {\em Proceedings of the 2019 3rd international conference on
  biometric engineering and applications}, pages 61--68, 2019.

\bibitem{lin2017focal}
Tsung-Yi Lin, Priya Goyal, Ross Girshick, Kaiming He, and Piotr Doll{\'a}r.
\newblock Focal loss for dense object detection.
\newblock In {\em Proceedings of the IEEE international conference on computer
  vision}, pages 2980--2988, 2017.

\bibitem{liu2021cross}
Ajian Liu, Xuan Li, Jun Wan, Yanyan Liang, Sergio Escalera, Hugo~Jair
  Escalante, Meysam Madadi, Yi Jin, Zhuoyuan Wu, Xiaogang Yu, et~al.
\newblock Cross-ethnicity face anti-spoofing recognition challenge: A review.
\newblock {\em IET Biometrics}, 10(1):24--43, 2021.

\bibitem{ijcai2022p165}
Ajian Liu and Yanyan Liang.
\newblock Ma-vit: Modality-agnostic vision transformers for face anti-spoofing.
\newblock In {\em Proceedings of the Thirty-First International Joint
  Conference on Artificial Intelligence, {IJCAI-22}}, pages 1180--1186.
  International Joint Conferences on Artificial Intelligence Organization,
  2022.

\bibitem{liu2021casia}
Ajian Liu, Zichang Tan, Jun Wan, Sergio Escalera, Guodong Guo, and Stan~Z Li.
\newblock Casia-surf cefa: A benchmark for multi-modal cross-ethnicity face
  anti-spoofing.
\newblock In {\em Proceedings of the IEEE/CVF Winter Conference on Applications
  of Computer Vision}, pages 1179--1187, 2021.

\bibitem{liu2021face}
Ajian Liu, Zichang Tan, Jun Wan, Yanyan Liang, Zhen Lei, Guodong Guo, and
  Stan~Z Li.
\newblock Face anti-spoofing via adversarial cross-modality translation.
\newblock {\em IEEE Transactions on Information Forensics and Security},
  16:2759--2772, 2021.

\bibitem{liu2019multi}
Ajian Liu, Jun Wan, Sergio Escalera, Hugo Jair~Escalante, Zichang Tan, Qi Yuan,
  Kai Wang, Chi Lin, Guodong Guo, Isabelle Guyon, et~al.
\newblock Multi-modal face anti-spoofing attack detection challenge at
  cvpr2019.
\newblock In {\em Proceedings of the IEEE/CVF Conference on Computer Vision and
  Pattern Recognition Workshops}, pages 0--0, 2019.

\bibitem{liu2022disentangling}
Ajian Liu, Jun Wan, Ning Jiang, Hongbin Wang, and Yanyan Liang.
\newblock Disentangling facial pose and appearance information for face
  anti-spoofing.
\newblock In {\em 2022 26th International Conference on Pattern Recognition
  (ICPR)}, pages 4537--4543. IEEE, 2022.

\bibitem{liu20213d}
Ajian Liu, Chenxu Zhao, Zitong Yu, Anyang Su, Xing Liu, Zijian Kong, Jun Wan,
  Sergio Escalera, Hugo~Jair Escalante, Zhen Lei, et~al.
\newblock 3d high-fidelity mask face presentation attack detection challenge.
\newblock In {\em ICCV Workshop}, pages 814--823, 2021.

\bibitem{liu2022contrastive}
Ajian Liu, Chenxu Zhao, Zitong Yu, Jun Wan, Anyang Su, Xing Liu, Zichang Tan,
  Sergio Escalera, Junliang Xing, Yanyan Liang, et~al.
\newblock Contrastive context-aware learning for 3d high-fidelity mask face
  presentation attack detection.
\newblock {\em IEEE Transactions on Information Forensics and Security},
  17:2497--2507, 2022.

\bibitem{liu20163d}
Siqi Liu, Pong~C Yuen, Shengping Zhang, and Guoying Zhao.
\newblock 3d mask face anti-spoofing with remote photoplethysmography.
\newblock In {\em European Conference on Computer Vision}, pages 85--100.
  Springer, 2016.

\bibitem{liu2018remote}
Si-Qi Liu, Xiangyuan Lan, and Pong~C Yuen.
\newblock Remote photoplethysmography correspondence feature for 3d mask face
  presentation attack detection.
\newblock In {\em Proceedings of the European Conference on Computer Vision
  (ECCV)}, pages 558--573, 2018.

\bibitem{liu2021multi}
Si-Qi Liu, Xiangyuan Lan, and Pong~C Yuen.
\newblock Multi-channel remote photoplethysmography correspondence feature for
  3d mask face presentation attack detection.
\newblock {\em IEEE Transactions on Information Forensics and Security},
  16:2683--2696, 2021.

\bibitem{Liu2018Learning}
Yaojie Liu, Amin Jourabloo, and Xiaoming Liu.
\newblock Learning deep models for face anti-spoofing: Binary or auxiliary
  supervision.
\newblock In {\em Proceedings of the IEEE conference on computer vision and
  pattern recognition}, pages 389--398, 2018.

\bibitem{liu2020disentangling}
Yaojie Liu, Joel Stehouwer, and Xiaoming Liu.
\newblock On disentangling spoof trace for generic face anti-spoofing.
\newblock In {\em European Conference on Computer Vision}, pages 406--422.
  Springer, 2020.

\bibitem{liu2021swin}
Ze Liu, Yutong Lin, Yue Cao, Han Hu, Yixuan Wei, Zheng Zhang, Stephen Lin, and
  Baining Guo.
\newblock Swin transformer: Hierarchical vision transformer using shifted
  windows.
\newblock In {\em Proceedings of the IEEE/CVF international conference on
  computer vision}, pages 10012--10022, 2021.

\bibitem{liu2022convnet}
Zhuang Liu, Hanzi Mao, Chao-Yuan Wu, Christoph Feichtenhofer, Trevor Darrell,
  and Saining Xie.
\newblock A convnet for the 2020s. arxiv e-prints (2022).
\newblock {\em arXiv preprint arXiv:2201.03545}, 2022.

\bibitem{manjani2017detecting}
Ishan Manjani, Snigdha Tariyal, Mayank Vatsa, Richa Singh, and Angshul
  Majumdar.
\newblock Detecting silicone mask-based presentation attack via deep dictionary
  learning.
\newblock {\em IEEE Transactions on Information Forensics and Security},
  12(7):1713--1723, 2017.

\bibitem{nesli2013spoofing}
Erdogmus Nesli and S{\'e}bastien Marcel.
\newblock Spoofing in 2d face recognition with 3d masks and anti-spoofing with
  kinect.
\newblock In {\em IEEE 6th International Conference on Biometrics: Theory,
  Applications and Systems (BTAS’13)}, pages 1--8, 2013.

\bibitem{shao2019multi}
Rui Shao, Xiangyuan Lan, Jiawei Li, and Pong~C Yuen.
\newblock Multi-adversarial discriminative deep domain generalization for face
  presentation attack detection.
\newblock In {\em Proceedings of the IEEE/CVF Conference on Computer Vision and
  Pattern Recognition}, pages 10023--10031, 2019.

\bibitem{steiner2016reliable}
Holger Steiner, Andreas Kolb, and Norbert Jung.
\newblock Reliable face anti-spoofing using multispectral swir imaging.
\newblock In {\em 2016 international conference on biometrics (ICB)}, pages
  1--8. IEEE, 2016.

\bibitem{szegedy2016rethinking}
Christian Szegedy, Vincent Vanhoucke, Sergey Ioffe, Jon Shlens, and Zbigniew
  Wojna.
\newblock Rethinking the inception architecture for computer vision.
\newblock In {\em Proceedings of the IEEE conference on computer vision and
  pattern recognition}, pages 2818--2826, 2016.

\bibitem{terhorst2020ser}
Philipp Terhorst, Jan~Niklas Kolf, Naser Damer, Florian Kirchbuchner, and Arjan
  Kuijper.
\newblock Ser-fiq: Unsupervised estimation of face image quality based on
  stochastic embedding robustness.
\newblock In {\em Proceedings of the IEEE/CVF conference on computer vision and
  pattern recognition}, pages 5651--5660, 2020.

\bibitem{wang2021towards}
Xintao Wang, Yu Li, Honglun Zhang, and Ying Shan.
\newblock Towards real-world blind face restoration with generative facial
  prior.
\newblock In {\em Proceedings of the IEEE/CVF Conference on Computer Vision and
  Pattern Recognition}, pages 9168--9178, 2021.

\bibitem{wang2018esrgan}
Xintao Wang, Ke Yu, Shixiang Wu, Jinjin Gu, Yihao Liu, Chao Dong, Yu Qiao, and
  Chen Change~Loy.
\newblock Esrgan: Enhanced super-resolution generative adversarial networks.
\newblock In {\em Proceedings of the European conference on computer vision
  (ECCV) workshops}, pages 0--0, 2018.

\bibitem{woo2023convnext}
Sanghyun Woo, Shoubhik Debnath, Ronghang Hu, Xinlei Chen, Zhuang Liu, In~So
  Kweon, and Saining Xie.
\newblock Convnext v2: Co-designing and scaling convnets with masked
  autoencoders.
\newblock {\em arXiv preprint arXiv:2301.00808}, 2023.

\bibitem{yang2021few}
Bowen Yang, Jing Zhang, Zhenfei Yin, and Jing Shao.
\newblock Few-shot domain expansion for face anti-spoofing.
\newblock {\em arXiv preprint arXiv:2106.14162}, 2021.

\bibitem{yu2020fas}
Zitong Yu, Jun Wan, Yunxiao Qin, Xiaobai Li, Stan~Z Li, and Guoying Zhao.
\newblock Nas-fas: Static-dynamic central difference network search for face
  anti-spoofing.
\newblock {\em IEEE transactions on pattern analysis and machine intelligence},
  43(9):3005--3023, 2020.

\bibitem{yu2020searching}
Zitong Yu, Chenxu Zhao, Zezheng Wang, Yunxiao Qin, Zhuo Su, Xiaobai Li, Feng
  Zhou, and Guoying Zhao.
\newblock Searching central difference convolutional networks for face
  anti-spoofing.
\newblock In {\em Proceedings of the IEEE/CVF Conference on Computer Vision and
  Pattern Recognition}, pages 5295--5305, 2020.

\bibitem{zhang2017mixup}
Hongyi Zhang, Moustapha Cisse, Yann~N Dauphin, and David Lopez-Paz.
\newblock mixup: Beyond empirical risk minimization.
\newblock {\em arXiv preprint arXiv:1710.09412}, 2017.

\bibitem{zhang2020face}
Ke-Yue Zhang, Taiping Yao, Jian Zhang, Ying Tai, Shouhong Ding, Jilin Li,
  Feiyue Huang, Haichuan Song, and Lizhuang Ma.
\newblock Face anti-spoofing via disentangled representation learning.
\newblock In {\em European Conference on Computer Vision}, pages 641--657.
  Springer, 2020.

\bibitem{zhang2020casia}
Shifeng Zhang, Ajian Liu, Jun Wan, Yanyan Liang, Guodong Guo, Sergio Escalera,
  Hugo~Jair Escalante, and Stan~Z Li.
\newblock Casia-surf: A large-scale multi-modal benchmark for face
  anti-spoofing.
\newblock {\em IEEE Transactions on Biometrics, Behavior, and Identity
  Science}, 2(2):182--193, 2020.

\bibitem{zhang2019dataset}
Shifeng Zhang, Xiaobo Wang, Ajian Liu, Chenxu Zhao, Jun Wan, Sergio Escalera,
  Hailin Shi, Zezheng Wang, and Stan~Z Li.
\newblock A dataset and benchmark for large-scale multi-modal face
  anti-spoofing.
\newblock In {\em CVPR}, 2019.

\bibitem{zhang2012face}
Zhiwei Zhang, Junjie Yan, Sifei Liu, Zhen Lei, Dong Yi, and Stan~Z Li.
\newblock A face antispoofing database with diverse attacks.
\newblock In {\em 2012 5th IAPR international conference on Biometrics (ICB)},
  pages 26--31. IEEE, 2012.

\bibitem{zhong2021sface}
Yaoyao Zhong, Weihong Deng, Jiani Hu, Dongyue Zhao, Xian Li, and Dongchao Wen.
\newblock Sface: Sigmoid-constrained hypersphere loss for robust face
  recognition.
\newblock {\em IEEE Transactions on Image Processing}, 30:2587--2598, 2021.

\end{thebibliography}
}

\end{document}